# Towards Emergent Language Symbolic Semantic Segmentation and Model Interpretability

Alberto Santamaria-Pang, James Kubricht, Aritra Chowdhury,
Chitresh Bhushan and Peter Tu

Artificial Intelligence, GE Research, Niskayuna, NY 12309, USA

**Abstract.** Recent advances in methods focused on the grounding problem have resulted in techniques that can be used to construct a symbolic language associated with a specific domain. Inspired by how humans communicate complex ideas through language, we developed a generalized Symbolic Semantic (S2) framework for interpretable segmentation. Unlike adversarial models (e.g., GANs), we explicitly model cooperation between two agents, a Sender and a Receiver, that must cooperate to achieve a common goal. The Sender receives information from a high layer of a segmentation network and generates a symbolic sentence derived from a categorical distribution. The Receiver obtains the symbolic sentences and cogenerates the segmentation mask. In order for the model to converge, the Sender and Receiver must learn to communicate using a private language. We apply our architecture to segment tumors in the TCGA dataset. A UNet-like architecture is used to generate input to the Sender network which produces a symbolic sentence, and a Receiver network cogenerates the segmentation mask based on the sentence. Our Segmentation framework achieved similar or better performance compared with state-of-the-art segmentation methods. In addition, our results suggest direct interpretation of the symbolic sentences to discriminate between normal and tumor tissue, tumor morphology, and other image characteristics.

## 1 Introduction

Current limitations in state-of-the-art Machine Learning (ML) and Artificial Intelligence (AI) include lack of interpretability and explainability; i.e., classical black-box approaches utilizing deep neural networks cannot provide evidence on how models behave. In medical applications, interpretability and explainability is a paramount requirement if we intend to rely on clinical diagnoses derived from automated systems. Inspired by the symbol grounding problem [1], the current work investigates synergies between deep learning Semantic Segmentation and Emergent Language (EL) models.

We further utilize general properties of EL architectures to facilitate model interpretability by demonstrating how black-box semantic segmentation can be extended to provide Symbolic Semantic ($S^2$) outputs. Corresponding sentences – drawn from a categorical distribution – are formed by integrating symbolic components into a conventional UNet-like architecture. We term this the Symbolic UNet (SUNet) framework.

Following description and analysis of the proposed framework, we explore the utility of symbolic segmentation masks towards direct data interpretability in clinical applications. Specifically, we utilize The Cancer Imaging Archive (TCGA) dataset to determine whether SUNet sentences correspond with meaningful semantics in neural imagery.

## 2     Literature Review

Interpretability and explainability of artificial intelligent (AI) systems is an important criterion for faster and wider adoption, especially in clinical applications [2]. Recently, several approaches for interpretable machine learning have been developed, with heavy emphasis on classification problems [3,4]. The majority of the existing approaches focus on explanation of the representation learned by the deep learning system through saliency maps, class activation maps, occlusion study, etc. [5,6,7,8,9]. Recent work from Natekar *et al.* [6] implemented a network dissection approach on a segmentation network to locate internal functional regions that identify human-understandable concepts like core and enhancing tumor regions. In a similar vein, Couteaux *et al.* [7] used activation maximization through gradient ascent, similar to DeepDream, to identify features in input images that the network is most sensitive to for segmentation of liver CT images. While these approaches do provide some interpretability of how networks represent data, the identified features or internal network components (layers, individual units, etc.) are not particularly interpretable by human experts [10, 11].

In this work we present an approach, inspired by the symbol grounding problem [1], that explicitly generates Symbolic Semantic (S2) and segmentation outputs. Our framework is inspired by Havrylov *et al.* [12], where multi-agent cooperation showed emergence of artificial language in natural images. In this approach, the sequence of symbols is modelled using paired Long Short-Term Memory (LSTM) networks. Cogswell *et al.* [13] also introduced compositional generality in emergent languages among multiple agents, similar to Larazidou *et al.* [14] who presented a series of studies investigating properties of language protocols generated by the agents exposed to symbolic and image data. In this work, we extend emergent language models to provide fully interpretable segmentation. To the best of our knowledge, no prior work has attempted to automatically express segmentation in contextually meaningful (symbolic) sentences. The two main innovations of this works are: i) emergent language extension to any segmentation architecture and ii) interpretation of symbolic expressions derived from segmentation tasks.

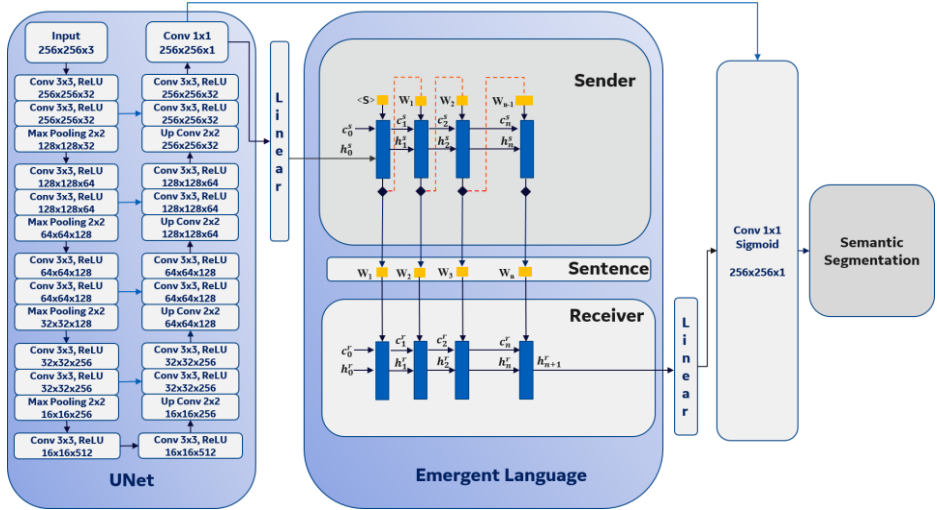

**Fig. 1.** Schematic of the proposed SUNet architecture composed of a UNet and Emergent Language (EL) network. Both networks are encouraged to cooperate in order to generate an appropriate segmentation mask. Once cooperation is achieved, semantic meaning of EL sentences are generated. These sentences are notably interpretable in the context of semantic segmentation output.

## 3   Methods

We present our $S^2$-Segmentation framework for simultaneous segmentation and emergent language generation. In general, we assume the following:

1. There is a segmentation network that provides a segmentation output $x$.
2. There is a vocabulary V = $\{w_1, w_2, …, w_{N_V}\}$. A sentence $S_{N_w}$ of length $N_w$ is a sequence of words $\{w_1, w_2, …, w_{N_w}\}$.
3. A Sender agent or network which receives the segmentation output $x$ and generates a sentence $S_{N_w}$ of length $N_w$, where $S_{N_w}$ = $Sender(x)$.
4. A Receiver agent or network, which obtains the symbolic sentence $S_{N_w}$ and generates an output $x'$ = $Receiver(S_{N_w})$.
5. The final segmentation is co-generated from: $x$ and $x'$.

To demonstrate applicability, we use a UNet network as implemented in [15], we omit the last sigmoid function (Fig 1, left side) to generate an output $x$. We include an Emergent Language (EL) network to generate a second output $x'$. Segmentation is obtained by concatenating $x$ and $x'$ and applying a sigmoid function (Fig. 1, right side). We train from end-to-end the segmentation and EL networks and when both converge, we conclude that an interpretable symbolic language has emerged. Next, we outline how to co-train both networks while reusing the original segmentation loss function.

## 3.1 Sender and Receiver Network

We implemented a variant of the Sender and Receiver networks reported in Havrylov *et al.* [12] using stacked LSTM models; see Hochreiter [16].

**Sender Network.** Input is: i) a tensor $x$ from the last convolutional layer output from the UNet, ii) a token <S> representing the start of the message. First, a linear transformation **Linear($x$)** is applied and passed to the stacked LSTM network. The initial hidden and cell state $h_0^s$ and $c_0^s$, respectively are set to zero. Unlike a conventional LSTM, the implemented LSTM samples a single symbol from a categorical distribution $w \sim \mathbf{Cat}(p_v^n)$, where $p_v^n$ represents the class probabilities with respect to the symbols in the vocabulary V at iteration $n$. Given that we are sampling from a categorical distribution, it is evident that this operation is not differentiable and therefore we cannot estimate the gradient during back propagation. To estimate a gradient during training, the Gumbel-Softmax trick [17] for discrete variables is implemented. Then, at each iteration $n$, we estimate a single symbol or word $w_i$ as follows:

$$w_i = G_\tau(p_i^n) = \frac{(\exp(\log(p_i^n) + g_i))/\tau)}{\sum_{j=1}^{v}(\exp(\log(p_j^n) + g_j))/\tau)} \quad (1)$$

where $\tau$ is the temperature parameter that regulates the Gumbel-Soft-max operator $G_\tau$ (diamond icons in Fig 1) and $g_i$ is Gumbel(0,1). The output of the Sender is the last hidden state $h_{n+1}^s$ which encodes the sentence as a sequence of words $w_i$ as: $h_{n+1}^s = \text{LSTM}(w_i, h_n^s, c_n^s)$. During inference, we do not apply the Gumbel-Soft-max operator [17], making $h_{n+1}^s$ fully deterministic. Then, the generated sentence is represented as $S_{N_w} = Sender(x)$.

**Receiver Network.** Unlike the Sender, the Receiver is implemented as a standard LSTM model. Input to the Receiver is the Sender's last hidden state (which encodes the sentence $S_{N_w}$). The initial hidden and cell states $h_0^r$ and $c_0^r$ respectively are set to zero. We then apply a linear transform **Linear($h_{n+1}^r$)** to the Receiver's last hidden state. The output of the receiver is $x' = Receiver(S_{N_w})$. During this process, the Sender and Receiver are encouraged to establish a communication protocol and if successful we conclude that a new emergent language has been produced. In order to generate a deterministic output, during inference we encode the categorical variable as a one-hot vector.

**Semantic Symbolic Segmentation:** To generate the final segmentation, we concatenate $x$ and $x'$ as: **Concat($x, x'$)**. We then apply a Convolution operator followed by batch normalization to produce a tensor of the same dimensions as $x$. The final segmentation output is obtained by applying a Sigmoid function. It is worth mentioning that this method is generally applicable to any segmentation network with the original loss function. Mathematically, the Emergent Language model can be interpreted as a regularization network, which forces the original segmentation network to have an internal feedback and control mechanism. In our experiments we noticed an average of six percent increase in performance.

## 4    Results

Imaging data was obtained from The Cancer Imaging Archive (TCGA) dataset. We analyzed 110 subjects as described in Buda *et al.* [15]. We integrated our Emergent Language module as described in the previous section. We associate each symbol to an integer number, and a segmentation mask is generated after applying a single connected component step, as reported in [15]. Table 1 presents a performance comparison between the UNet model from [15] and our SUNet model. For the SUNet, different sentence length $N_W$ and vocabulary size $N_V$ was selected. The size of hidden dimension tensor and cell state tensor was set to 1024 and 2048 respectively (Sender and Receiver). We found best performance with vocabulary corresponding to a sentence length of 10 and vocabulary of 10K symbols. In Figure 2, columns, w0 to w10, show a visualization of emergent symbols (per 2D section) and the last four columns correspond to area and eccentricity, tumor detection and ground truth. We observe that different patterns of symbols emerge if the 2D image corresponds to normal or tumor tissue. For example, for subject CS 4941 (left), the regular expressions associated with normal tissue are i) **657, 653, \***; ii) **657, 8927, 3330, \***; iii) **657, 3785, \***; whereas the expression associated with tumor presence are i) **8584, \***; ii) **1168, \***; iii) **3912, \***. Unlike the last symbols in the sentence, we notice that the initial symbols appear to best characterize tissue type. This is expected given that the LSTM model is hierarchical.

**Table 1.** Comparative performance from [15] and our proposed SUNet framework.

| Prediction | Buda *et al.* [15] | $N_w = 4$<br>$N_V = 100$ | $N_w = 10$<br>$N_V = 10k$ | $N_w = 10$<br>$N_V = 50k$ | $N_w = 20$<br>$N_V = 10k$ | $N_w = 20$<br>$N_V = 1k$ |
|---|---|---|---|---|---|---|
| Best validation (mean DSC) | 81.4 | 89.1 | **90.0** | 89.3 | 81.5 | 78.8 |

**Fig. 2.** Example of emergent language sentence across slides for two subjects and for model corresponding to $N_w = 10$ and $N_V = 10k$. Columns w1 through w10 show distribution of symbols. The last four columns correspond to area, eccentricity, tumor detection and ground truth.

## 4.1 Regression Performance

Results in the previous sections indicate that symbolic expressions can be used to predict segmentation masks on a fluid-attenuated inversion recovery (FLAIR) brain images, and those symbols are informative with respect to presence or absence of tumor. Further analyses were conducted to determine whether individual symbols in emergent language expressions can predict tumor presence (Tumor), morphology (Area, Eccentricity), location (Laterality, Location) and patient genome (miRNA). Specifically, we examined which expression symbol (i.e., first, second, etc.) is best at predicting each outcome for all candidate models. Results from this analysis are reported in Table 2.

**Table 2.** Linear and logistic regression results. The top row indicates model parameters, where $N_W$ is the sentence length, and $N_V$ is the vocabulary size. The following rows indicate prediction type, and columns indicate which model was most predictive of the outcome, and which specific symbol $S^*$ performed best. Squared Pearson correlation coefficients are provided for continuous variables (Area, Eccentricity), and McFadden's pseudo-$R^2$ is reported for categorical outcomes (Tumor, Laterality, Location, Histology Type, Histology Grade and miRNA).

| Prediction | $N_w = 4$ $N_V = 100$ | | $N_w = 10$ $N_V = 10k$ | | $N_w = 10$ $N_V = 50k$ | | $N_w = 20$ $N_V = 10k$ | | $N_w = 20$ $N_V = 1k$ | |
|---|---|---|---|---|---|---|---|---|---|---|
| | $S^*$ | $R^2$ | $S^*$ | $R^2$ | $S^*$ | $R^2$ | $S^*$ | $R^2$ | $S^*$ | $R^2$ |
| Tumor | $S_4$ | 0.27 | $S_1$ | 0.48 | $S_1$ | 0.29 | $S_4$ | 0.59 | **$S_1$** | **0.67** |
| Area | $S_3$ | 0.27 | $S_2$ | 0.43 | $S_1$ | 0.33 | **$S_1$** | **0.44** | $S_2$ | 0.39 |
| Eccentricity | $S_4$ | 0.06 | $S_2$ | 0.11 | $S_1$ | 0.10 | $S_1$ | 0.15 | **$S_5$** | **0.19** |
| Laterality | $S_3$ | 0.22 | $S_1$ | 0.37 | $S_4$ | 0.22 | **$S_1$** | **0.41** | $S_2$ | 0.33 |
| Location | $S_4$ | 0.11 | **$S_1$** | **0.23** | $S_3$ | 0.12 | $S_4$ | 0.19 | $S_2$ | 0.16 |
| Histology Type | $S_3$ | 0.06 | **$S_2$** | **0.12** | $S_4$ | 0.06 | $S_1$ | 0.10 | $S_2$ | 0.08 |
| Histology Grade | $S_1$ | 0.03 | **$S_1$** | **0.09** | $S_1$ | 0.06 | $S_2$ | 0.09 | $S_2$ | 0.08 |
| miRNA | $S_3$ | 0.07 | **$S_1$** | **0.13** | $S_1$ | 0.06 | $S_1$ | 0.13 | $S_1$ | 0.09 |

The statistical model used to predict each outcome varied. For binary data (Tumor), a binary logistic regression model was used. For outcomes with multiple classes (Laterality, Location, miRNA), a multinomial logistic regression was performed. Finally, linear regression was performed on continuous data (Area, Eccentricity). Both multinomial logistic and linear regression were performed on data where a tumor was present. We report squared Pearson correlation coefficient $R^2$ values for continuous outcomes and McFadden's pseudo-$R^2$ for categorical outcomes. Results indicate high correlations between expression symbols and tumor presence, area and laterality. Both tumor eccentricity and location were moderately correlated with emergent language symbols, although histology type, histology grade and patient genome (miRNA) achieved lesser correlations. When only four symbols were used ($V = 4$), expressions were least predictive of each outcome, as indicated by correspondingly lower correlations.

It is worth noting, however, that symbols occurring later in each expression ($S_3$ and $S_4$) were most informative with respect to each outcome, whereas the opposite trend was observed in the remaining models. With the exception of tumor eccentricity, all outcomes were best explained by either the first or second symbol in emergent language expressions ($S_1$ or $S_2$). This suggests that while symbols occurring later in the expression are important for capturing and refining semantics, the weight of explainability appears to fall on earlier elements.

**Interpretability and Explainability.** Figure 3 shows a visualization of symbol interpretability and quality of segmentation in the analyzed 2D images. We show monochromatic images after mapping from three color channels (FLAIR images) to grayscale. Green and red contours represent ground truth and predicted tumor contours, respectively. Predictions correspond to our SUNet model with 10K symbols and sentence length 10. Post-processing was applied as reported in [15].

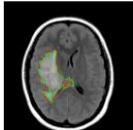
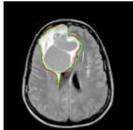
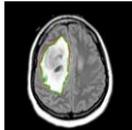
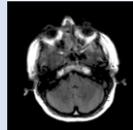
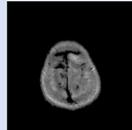
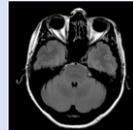
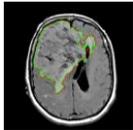
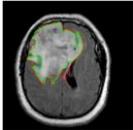
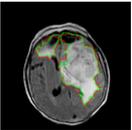
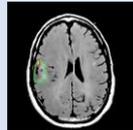
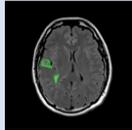
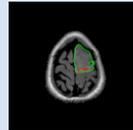
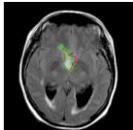
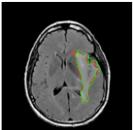
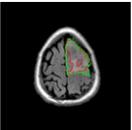
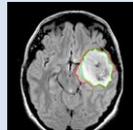
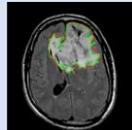
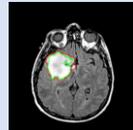

**Fig. 3.** Examples of brain images associated with (top) tumor presence, (middle) tumor size or area and (bottom) tumor eccentricity. Subject identifiers are reported in bold text, followed by the associated image slice. The first two emergent language symbols ($S_1$, $S_2$) are reported, followed by * indicating "all remaining symbols".

From top to bottom we show images for: i) tumor present or absent, ii) tumor area and iii) tumor eccentricity across several representative subjects that correspond to the first two symbols within the generated ten symbol sentence. The top rows show tissue characterization for tumor or no tumor. The top left shows the segmented tumors along with the interpretable symbolic regular expressions: **357, 6563, 3956, \***; **5886, 4043, 5114, \*** and **5886, 3101, 5114, \***, where the \* indicates any remaining symbol in the sentence. We also notice that while we can generate a symbolic sentence, the quality of the segmentation highly matches the ground truth. This suggests that the Receiver network was able to: i) interpret correctly the sentence generated by the Sender network and ii) co-generate correctly the segmentation mask from the full sentence. The top-right row suggests a trend from the regular expressions: **657, 653, 6131, \***; **657, 3785, 6613, \*** and **657, 653, 6531, \*** for bottom (slice 7), top (slice 29) and bottom (slice 7) areas of the brain respectively. It should be noted that $R^2$ value for the predictive model was 0.48 (high confidence). The middle rows correspond to area. On the left and right show examples of large and small tumor. For example, symbols **7313, 4626, 8872, \***; **8584, 4043, 5114, \*** and **6670, 3578, 252, \*** may be associated with large tumors, whereas **657, 2863, 2863, \***; **657, 653, 6613, \*** and **657, 653, 6613, \*** may be indicative of small tumors. Here the $R^2$ value was 0.43 (high confidence). The bottom row shows an example of prediction with medium-low confidence ($R^2 = 0.11$) for small tumors with high eccentricity and large tumors with high eccentricity.

## 5 Conclusions and Future Work

We have presented a generalized framework to enable Symbolic Semantic ($S^2$) Segmentation in medical imagery. Unlike standard semantic segmentation, our proposed $S^2$-Segmentation uses an Emergent Language model via Sender and Receiver agents to produce symbolic sentences. Such symbolic sentences are used by the Receiver agent to co-generate the final segmentation mask. The proposed framework can be applied to any segmentation network allowing direct interpretation of predictions. While integer symbols are not particularly intuitive, future work will build on recent success translating symbols to natural language through neural machine translation architectures. To the best of our knowledge, this is the first work to demonstrate feasibility of Emergent Language models towards semantic segmentation interpretation and explanation.

We implemented a Symbolic UNet (SUNet) architecture for tumor segmentation and interpretation using the TCGA dataset. Statistical analysis suggests feasibility in associating symbolic sentences with clinically relevant information, such as tissue type (tumor vs normal), object morphology (area, eccentricity), object localization (tumor laterality and location), tumor histology and genomics data. For future work we plan to extend validation to larger datasets and investigate different network architectures. Similarly, we plan carry on a more detailed analysis on the possibility of using symbolic expressions to drive interpretability in complex bioinformatics tasks towards personalized diagnostics and precision medicine linked with pathology data.

**References**

1. Harnad, S.: Minds, Machines and Turing: The Indistinguishability of Indistinguishables. Journal of Logic. Language, and Information 9(4): 425–445, (2000).
2. Kelly, C.J., Karthikesalingam, A., Suleyman, M. *et al.*: Key challenges for delivering clinical impact with artificial intelligence. BMC Med 17, 195 (2019).
3. Doshi-Velez, F., Kim, B.: Towards a rigorous science of interpretable machine learning. arXiv preprint arXiv:1702.08608 (2017).
4. Adadi, A., Berrada, M.: Peeking inside the black-box: A survey on explainable artificial intelligence (XAI). IEEE Access 6, 52138–52160 (2018).
5. Gilpin, L. H., Bau, D., Yuan, B. Z., Bajwa, A., Specter, M., Kagal, L.: Explaining Explanations: An Overview of Interpretability of Machine Learning. IEEE 5th International Conference on Data Science and Advanced Analytics (DSAA), Italy, pp. 80-89, (2018).
6. Natekar, P., Kori, A., Krishnamurthi, G.: Demystifying Brain Tumor Segmentation Networks: Interpretability and Uncertainty Analysis. Frontiers in Computational Neuroscience, 14, 6, (2020).
7. Couteaux, V., Nempont, O., Pizaine, G., Bloch, I.: Towards Interpretability of Segmentation Networks by Analyzing DeepDreams. In Interpretability of Machine Intelligence in Medical Image Computing and Multimodal Learning for Clinical Decision Support (pp. 56-63). Springer, Cham., (2019).
8. Ribeiro, M. T., Singh, S., Guestrin, C.: "Why should I trust you?" Explaining the predictions of any classifier. In Proceedings of the 22nd ACM SIGKDD international conference on knowledge discovery and data mining (pp. 1135-1144), (2016).
9. Simonyan, K., *et al*.: Deep Inside Convolutional Networks: Visualising Image Classification Models and Saliency Maps. arXiv 1312.6034.
10. Yeh, C. K., Hsieh, C. Y., Suggala, A. S., Inouye, D., Ravikumar, P.: How Sensitive are Sensitivity-Based Explanations?. arXiv preprint arXiv:1901.09392, (2019).
11. Kindermans, P. J., Hooker, S., Adebayo, J., Alber, M., Schütt, K. T., Dähne, S., Erhan, D., Kim, B.: The (un) reliability of saliency methods. In Explainable AI: Interpreting, Explaining and Visualizing Deep Learning (pp. 267-280). Springer, Cham, (2019).
12. Havrylov, S., Tritov, I.: Emergence of Language with Multi-Agent Games: Learning to Communicate with Sequences of Symbols. Advances in Neural Information Processing Systems, Vol 30, pp. 2149-2159, (2017).
13. Cogswell, Lu, M., J., Lee, S., Parikh, D., Batra, D.: Emergence of Compositional Language with Deep Generational Transmission. *arXiv preprint arXiv:1904.09067*, April, (2019).
14. Lazaridou, A., Herman, K. M., Tuyls, K., Clark, S.: Emergence of linguistic communication from referential games with symbolic and pixel input. *arXiv preprint arXiv:1804.03984*, April, (2018).
15. Buda, M., Saha, A., Mazurowski, M. A.: Association of genomic subtypes of lower-grade gliomas with shape features automatically extracted by a deep learning algorithm. Computers in Biology and Medicine, (2019).
16. Hochreiter, S., Schmidhuber, J.: LSTM can solve hard long time lag problems. In Advances in neural information processing systems 9. Cambridge, MA: MIT Press, (1997).
17. Jang, E., Gu, S., and Poole, B.: Categorical reparameterization with Gumbel-Softmax. arXiv preprint arXiv:1611.01144, (2016).